\pdfoutput=1

\documentclass[11pt]{article}

\usepackage[preprint]{acl}
\usepackage{booktabs}

\usepackage{times}
\usepackage{latexsym}

\usepackage{xcolor}
\usepackage{graphicx}
\usepackage{multirow}
\usepackage[T1]{fontenc}

\usepackage{amsmath}

\usepackage[utf8]{inputenc}

\usepackage{microtype}

\usepackage{inconsolata}

\title{Confabulation:\\ The Surprising Value of Large Language Model Hallucinations}

\author{Peiqi Sui$^{1*}$\quad Eamon Duede$^{2\dag}$\quad Sophie Wu$^{1\triangle}$\quad Richard Jean So$^{1\ddagger}$ \\
$^1$McGill University, CA\\
$^2$Harvard University, USA\\
\texttt{$^*$peiqi.sui@mail.mcgill.ca}\quad \texttt{$^\dag$eduede@g.harvard.edu}\quad\\ 
\texttt{$^\triangle$sophie.wu@mail.mcgill.ca}\quad
\texttt{$^\ddagger$richard.so@mcgill.ca}\\
\\
Forthcoming at 62nd Annual Meeting of the Association for Computational Linguistics\\ (ACL2024 Main Conference)
}

\begin{document}
\maketitle
\begin{abstract}
This paper presents a systematic defense of large language model (LLM) hallucinations or `confabulations' as a potential resource instead of a categorically negative pitfall. The standard view is that confabulations are inherently problematic and AI research should eliminate this flaw. In this paper, we argue and empirically demonstrate that measurable semantic characteristics of LLM confabulations mirror a human propensity to utilize increased narrativity as a cognitive resource for sense-making and communication. In other words, it has potential value. Specifically, we analyze popular hallucination benchmarks and reveal that hallucinated outputs display increased levels of narrativity and semantic coherence relative to veridical outputs. This finding reveals a tension in our usually dismissive understandings of confabulation. It suggests, counter-intuitively, that the tendency for LLMs to confabulate may be intimately associated with a positive capacity for coherent narrative-text generation.

\end{abstract}

\section{Background}

Artificial intelligence is rapidly becoming ubiquitous in science and scholarship with autoregressive large language models (LLMs) leading the way \cite{duede2024oil}. Since the concept's integration into natural language processing (NLP) discourse, hallucination in LLMs has been widely viewed as a conspicuous social harm and a critical bottleneck to real-world applications of LLMs. Both popular, comprehensive academic surveys \cite{ji2023survey,kaddour2023challenges} and public-facing technical reports \cite{OpenAI2023gpt,touvron2023llama,Google023palm} position the problem of hallucination as one of LLM's chief ethical and safety pitfalls, one that should be heavily mitigated in conjunction with other concerns like bias and toxicity. Yet, despite a wide range of such mitigation efforts \cite{huang2023survey}, hallucinations perdure, presenting an imminent risk to model trustworthiness \cite{luo2024hallucination}, with epistemic consequences in truth-sensitive domains like law \cite{curran2023hallucination}, medicine \cite{pal-etal-2023-med}, finance \cite{kang2023deficiency}, science \cite{alkaissi2023artificial,duede2022instruments,duede2023deep}, and education \cite{zhou2024teachers}.

\begin{table*}[ht!]
\centering
\begin{tabular}{p{3.85cm} p{11.15cm}}
\toprule
\textbf{Type} & \textbf{Definition of Confabulation} \\
\midrule
Clinical Def. & ``A neuropsychiatric disorder wherein a patient generates a false memory without the intention of deceit'' \cite{PMID:30725646} \\
& \\
NLP Def. & ``...where an LLM seemingly fills gaps in the information contained in the model with plausibly-sounding words'' \cite{nejjar2023llms}\\
& \\
(Clinically-Informed) NLP Def. & ``The generation of narrative details that, while incorrect, are not recognized as such... mistaken reconstructions of information which are influenced by existing knowledge, experiences, expectations, and context'' \cite{smith2023hallucination}\\

\bottomrule
\end{tabular}
\caption{Extant definitions of `confabulation' from the literature.}
\label{definitions}
\end{table*}

With so much at stake, NLP researchers have naturally developed a \textbf{normative} stance against hallucinations, rightly extending epistemic worries to more far-reaching, ethical concerns like misinformation \cite{li2023dark}, deception \cite{zhan2023deceptive}, fairness \cite{wang2023human}, and the broader goals of alignment research \cite{ouyang2022training}. Consequently, the observed commitment to uniform reduction of hallucinations to negligible levels is not just seen as a technical challenge but also a critical component of the broader mission to mitigate the societal stigmas and systemic risks associated with the broad deployment and wide adoption of LLMs.

However, a smaller body of work advances the view that hallucinations are not inherently harmful. According to this view, hallucinated outputs have particular affordances. For instance, such outputs can be utilized for the exploration of domain-specific scenarios where properties of hallucination could be leveraged constructively \cite{duede2024humanities}. \citet{cao-etal-2022-hallucinated} demonstrate that certain types of hallucinations are factual and could positively contribute to text summarization. \citet{yao2023llm} explore the potential of hallucinatory responses as adversarial examples for enhancing the robustness of models. Following an existing practice in computer vision to use hallucinated materials as synthetic training data, \cite{fei-etal-2023-scene} devise a visual scene hallucination mechanism to dynamically fill in the missing visual modality for text-only inference with multimodal machine translation models \cite{mckee2021multi,wu2023hallucination,shah2023halp}.

This \textbf{explorative} view highlights the potential affordances and plausible necessity of hallucinations. The possible necessity of hallucinations is supported by recent research positing that hallucinations are a statistical inevitability \cite{kalai2023calibrated,xu2024hallucination}, 
and impossible to eliminate from LLMs as efforts to do so would be limited to a trade-off between generativity, creativity. and information accuracy \cite{lee2023mathematical,sinha2023mathematical,zhang2023user}. 
Moreover, in many domain-specific applications, achieving an optimized balance between creativity and factuality is more effective for maximizing the utility of LLMs than merely attempting to eliminate hallucinations. LLM use cases where hallucinations could be particularly valuable include discovering novel proteins \cite{anishchenko2021novo}, providing inspiration for creative writing \cite{mukherjee2023creative}, and formulating innovative legal analogies \cite{dahl2024large}.

In this paper, we attempt to broaden the concept of hallucination and argue that hallucination is closer in kind to the concept of `confabulation', a term that has already gained popularity in public discourse on AI but has yet to become widespread in the academic literature. Reorienting our conception of what is typically conceived of as hallucination towards confabulation offers a more flexible way to characterize, measure, and analyze factually inaccurate outputs. 

We first operationalize a common measure of narrativity following \cite{piper2021narrative,piper2022toward,antoniak2023people} and then demonstrate empirically (see Section~\ref{sec:results}) that `hallucinations' manifest significantly higher `narrativity' than strictly factual outputs. Scholarship in cognitive science and cultural analytics suggests that narrativity provides beneficial cognitive and social effects and that, by extension, confabulations that demonstrate such features are likely to offer similar affordances. Taking this broader perspective, in turn, allows researchers to move beyond a strictly normative approach that narrowly focuses on identifying and discarding hallucinations. As such, casting hallucination as confabulation offers a practical and tangible framework for adjusting to the nuance of hallucinations. 

\section{Related Work}
\subsection{Confabulation vs Hallucination}
While both are anthropomorphizing analogies borrowed from psychiatry, `confabulation' has recently emerged as a preferred alternative to `hallucination' in the public discourse on AI for three reasons. First, confabulation does not pertain to precepts or stimuli, thus avoiding the thorny implication that LLMs have sensory experiences or even consciousness \cite{smith2023hallucination}. Second, as a more neutral term, confabulation is more sensitive to groups dealing with mental health disorders and better at signaling a lack of malicious intent in that the output is not to be thought of as a fabrication \cite{emsley2023chatgpt}. Finally, the concept's applicability to non-pathological scenarios is supported by research from psychiatry suggesting that everyday memory reconstruction often involves some degree of confabulation \cite{french2009false} and is equally common for healthy individuals to inadvertently fictionalize details of stories without the intention to deceive \cite{riesthuis2023factors} (points we return to in detail in Section~\ref{sec:defense}).

Despite conceptual distinction and advantages over `hallucination', `confabulation' is still commonly used interchangeably with hallucination in the NLP literature \cite{romera2023mathematical}, quite often trivially so \cite{liu2023correction,zhan2023deceptive,suraworachet2024predicting}. From the perspective of model architecture, confabulation has been considered as a formal disparity in confidence level between a model's queried output and probed internal representation of truthfulness \cite{liu2023cognitive}. More severe implications of the term have been discussed from the perspective of medicine. For instance \cite{hatem2023chatbot, brender2023chatbot} have debated the trade-off between the stigmatization that attaches to `hallucination' and the semantic baggage that attends the anthropomorphism of LLM outputs characterized as confabulation. However, neither can offer a more substantive definition than their cited clinical baselines \cite{PMID:30725646}.

\subsection{Towards a Narrative-Centered Definition of Confabulation}

However, extant definitions (see Table~\ref{definitions}) in the literature do not sufficiently account for confabulation's social and cognitive benefits in human communication. That is, current definitions attributed to LLMs fail to consider that, unlike hallucination, confabulation affords communicative and cognitive benefits to humans when endeavoring to fill in gaps in their knowledge of contextually relevant details to produce coherent linguistic communication. A conflation of confabulation with hallucination limits the potential affordances to NLP research that the concept of confabulation could play. 

To address this, we define confabulation in a Narrative-Centric way as a latent narrative impulse to generate more substantive and coherent outputs ---a characteristic of LLM textual outputs that closely mirrors the human predisposition to storytelling as a cognitive resource for sensemaking. Specifically, confabulation is a narrative impulse to schematize the information at hand into self-consistent stories, even if there might not be enough available details to do so, in which case would result in the generation of fictional yet plausible information. When humans do not have access to sufficient information to formulate coherent semantic meaning, they often confabulate to `fill in the blanks' with self-consistent narratives that are not necessarily factual but bear close semantic verisimilitude to reality. Following existing conceptions of narrativity in narrative understanding \cite{piper2021narrative}, we further hypothesize that the extent of this compensatory storytelling can be measured as a scalar construct that quantifies the degree of narrativity. 

\begin{table*}[ht]
\centering
\begin{tabular}{lccccccccc}
\toprule
& \multicolumn{3}{c}{\textbf{\texttt{FaithDial}}} & \multicolumn{2}{c}{\textbf{\texttt{HaluEval}}} & \multicolumn{3}{c}{\textbf{\texttt{BEGIN}}} \\
\cmidrule(lr){2-4} \cmidrule(lr){5-6} \cmidrule(lr){7-9}
& Hallucinated & Partial & Truth & Hallucinated & Truth & Hallucinated & Partial & Truth \\
\midrule
\textbf{Count} & 4485 & 14108 & 2852 & 10000 & 10000 & 1019 & 239 & 1139 \\
\textbf{Mean} & \textbf{0.620} & 0.606 & 0.518 & \textbf{0.655} & 0.638 & \textbf{0.658} & 0.612 & 0.561 \\
\textbf{Std} & 0.178 & 0.178 & 0.186 & 0.138 & 0.168 & 0.183 & 0.188 & 0.187 \\
\textbf{Min} & 0.041 & 0.052 & 0.061 & 0.167 & 0.029 & 0.064 & 0.073 & 0.059 \\
\textbf{25\%} & 0.500 & 0.486 & 0.379 & 0.563 & 0.528 & 0.541 & 0.477 & 0.433 \\
\textbf{50\%} & 0.640 & 0.624 & 0.524 & 0.669 & 0.664 & 0.695 & 0.642 & 0.567 \\
\textbf{75\%} & 0.756 & 0.741 & 0.664 & 0.759 & 0.768 & 0.799 & 0.761 & 0.698 \\
\textbf{Max} & 0.975 & 0.974 & 0.959 & 0.952 & 0.985 & 0.965 & 0.956 & 0.965 \\
\bottomrule
\end{tabular}
\caption{Summary statistics for \texttt{Narrativity} across our three benchmark datasets. We observe that, in all three datasets, outputs labeled as `hallucinated' text have, on average, the highest narrativity.}
\label{tab:summary_statistics}
\end{table*}

\section{Data, Methods, and Results}
\label{sec:results}
\subsection{Empirical Results for Higher Narrativity in Hallucinations}

To validate the narrative-rich properties of confabulation, we compare the narrativity of hallucinations and their ground truth counterparts across three popular dialogic hallucination benchmarks:

\begin{itemize}
    \item \texttt{FaithDial} \cite{dziri-etal-2022-faithdial} is a hallucination-free dialogue benchmark between an information-seeking user and a chatbot, adapted from the Wizard of Wikipedia (WoW) benchmark \cite{dinan2018wizard}. Mechanical Turk annotators labeled WoW's human-generated responses as either `Hallucination' or as truthful responses. Truthful responses are broken into three classes: `Entailment', `Uncooperative', and `Generic', and produced faithful and knowledge-grounded edits to 21,445 original responses, all of which we sample for our study.

    \item \texttt{BEGIN} is a preliminary study of \texttt{FaithDial} conducted to select an existing benchmark for subsequent large-scale annotation and editing \cite{dziri-etal-2022-origin}. As a smaller expert-curated set, it consists of information-seeking queries, as well as both human-written and model-generated (GPT-2, DoHA, and CRTL) responses, each labeled with a slightly different hallucination taxonomy than \texttt{FaithDial} (with the addition of `Partial Hallucination' as a label) by expert annotators. We adopt \texttt{BEGIN} as a model and dataset agnostic validation of our findings on \texttt{HaluEval}, to confirm the consistency and robustness the narrative patterns across different datasets and models.

    \item \texttt{HaluEval} \cite{li-etal-2023-halueval} is a comprehensive dataset featuring plausible but hallucinated ChatGPT generations alongside their groundtruth counterparts. Instead of the more fine-grained hallucination labels of \texttt{FaithDial} and \texttt{BEGIN}, \texttt{HaluEval} only differentiates between hallucinated and ground truth responses. We only utilize the dialogue sections of \texttt{HaluEval} with 10,000 samples to maintain domain consistency with the other benchmarks.
\end{itemize}

\begin{figure*}
\centering
\includegraphics[width=\linewidth]{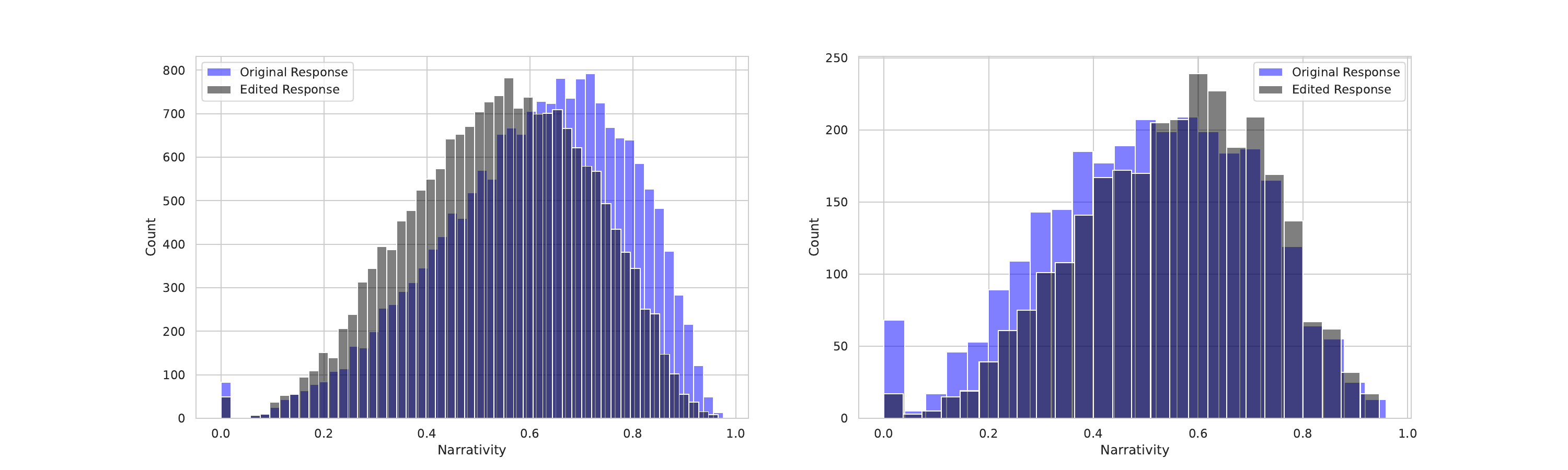}
\caption{The left panel illustrates distribution for narrative score of hallucinated outputs (blue) and the edited version of the output (gray) in the \texttt{FaithDial} dataset. The hallucinated texts are, in general more narrative rich than those that are edited to resolve inaccuracies. The right panel illustrates distribution for non-hallucinated texts from the \texttt{FaithDial} dataset.}
\label{fig:faith_hallucination}
\end{figure*}

For both the \texttt{FaithDial} and \texttt{BEGIN} datasets, we treat all outputs that do not contain a `hallucination' label as `truth' and all outputs that contain the `hallucination' label as well as an additional truthful label as a `partial' hallucination/truth. This aggregation allows for more direct comparison across datasets. In Table \ref{tab:summary_statistics} and Figure \ref{fig:faith_hallucination}, we present empirical support for the claim that confabulated texts, in general, exhibit increased levels of narrativity and, as such, can be viewed as a form of narrative-rich behavior.

\textbf{Modeling Narrativity}. The degree of narrativity is measured as the softmax probability output of an ELECTRA-large based \cite{clark2020electra} text-classification model fine-tuned on an expert-annotated narrative detection dataset recently curated from Reddit \cite{antoniak2023people}. Antoniak et al. observe that a fine-tuned RoBERTa model outperforms zero-shot and few-shot approaches with larger models like GPT-4 at distinguishing between texts that represent stories from texts that represent non-stories. Following this strategy, we fine-tune a series of encoder-only models for story detection as a text classification task, which perform similarly when evaluated against both Antoniak et al.'s bootstrapped test set (AUC=0.83-0.85) and Piper et al.'s larger (AUC=0.81-0.84) \cite{piper2022toward}. We then qualitatively evaluate each model's classifications to select the model (the fine-tuned ELECTRA-large) that best generalizes to the dialogue contexts of our target benchmarks and further verify its robustness through the manual evaluation of its inference output.

\textbf{Results.} Across all 3 benchmark datasets, we find that hallucinated \textit{qua} confabulated responses exhibit higher narrativity than both partial and non-hallucination categories and their own ground truth counterparts (See Table~\ref{tab:summary_statistics}). Additionally, we find significant correlations between the narrativity score and the hallucination label. In this study, we employed a binomial logistic regression model to investigate the predictive relationship between the independent variable narrativity $x$ and a binary dependent variable $y$ that indicates `hallucination' or `not hallucination' such that $P(y_i = 1 | x_i) = \frac{e^{\beta_0 + \beta_1 x_i}}{1 + e^{\beta_0 + \beta_1 x_i}}$. For the purposes of this exercise, we class partial hallucinations as 1. Give the slighly lower mean `narrativity' score for partial hallucination compared to pure hallucinated texts, we suspect that the results of our logistic regression, in fact, under-estimate the magnitude of the association. The logistic regression model allowed us to estimate the probability of $y$ being 1 as a function of $x$. The model's coefficients were estimated using maximum likelihood estimation (MLE). Table~\ref{tab:narrative_regs} indicates the results of fitting the model to predict whether an unedited response in FaithDial and BEGIN will likely be labeled as a hallucination. The positive coefficient and low p-value indicate that higher degrees of narrativity are a significantly predictive feature of the hallucination label in both benchmarks.

\begin{table}[!ht]
\centering
\begin{tabular}{lcc}
\hline
\textbf{Variable} & \textbf{Coefficient} & \textbf{Std. Error} \\
\hline
\texttt{Narrativity} & 0.631 $^{***}$ & 0.059 \\
\texttt{Intercept} & 0.368 $^{***}$ & 0.038 \\
\hline
\end{tabular}
\caption{Regression results of logistic regression. Dependent variable is the hallucination classification (1 for hallucinated output, 0 for ground truth output). Observations: 43,842 | Log Likelihood: -27,397.6 | $^{***}$ denotes significance at the 0.01 level.}
\label{tab:narrative_regs}
\end{table}

\section{In defense of confabulation}
\label{sec:defense}
In this section, we present a two-part argument that calls for a more careful and nuanced examination of confabulation and possible utilization of its affordances. We argue that confabulation's narrative-rich properties should not be viewed as a flaw but a hallmark for LLM alignment with a well-established human tendency to use narratives as a versatile tool for persuasion, identity construction, and social negotiation \cite{bruner1991narrative}. The normative view's unreflective dismissal of confabulation, in turn, would risk eliminating critical behavioral and cognitive capacities central to communication and sense-making from the capabilities of LLMs. We also consider how aspects of narrative theory could complicate and enrich our understanding of newly emergent concepts in NLP, like faithfulness and factuality.

\subsubsection{Empirical Support for Association of Narrativity and Coherence in Confabulated Texts}

To empirically verify the correlation between narrativity and coherence, we extend our analyses to include an evaluation of dialogic coherence by implementing DEAM, a state-of-the-art metric for the task \cite{ghazarian-etal-2022-deam}. DEAM measures coherence in dialogues with a RoBERTa-large model fine-tuned on conversation-level semantic perturbations designed to generate more natural samples of incoherence, which achieves better generalizability from their training set than other metrics for coherence evaluation. 

\begin{table}[!ht]
\centering
\begin{tabular}{lcc}
\hline
\textbf{Variable} & \textbf{Coefficient} & \textbf{Std. Error} \\
\hline
\texttt{Narrativity} & 0.372 $^{***}$ & 0.029 \\
\texttt{Intercept} & 0.433 $^{***}$ & 0.018 \\
\hline
\end{tabular}
\caption{Regression results of beta regression. Dependent variable is \texttt{Coherence} and indpendent variable is output \texttt{Narrativity}. Observations: 65,287 | Pseudo R$^2$: 0.004 | Log Likelihood: 109,935.2. | $^{***}$ denotes significance at the 0.01 level.}
\label{tab:coherence}
\end{table}

We find that higher narrativity is associated with higher coherence across all three benchmarks. Table~\ref{tab:coherence} shows the results of a beta regression with narrativity score as an independent variable and coherence as a dependent variable. Beta regression was used because all values in our dataset range between 0 and 1. Beta regression is particularly suited for modeling continuous variables restricted to the interval (0, 1), making it ideal for proportions and rates \cite{ferrari2004beta}. It accounts for the heteroscedasticity commonly observed in such data and provides more accurate estimates than linear regression in these contexts. Our regression model is simple: $\text{Beta}(\text{\texttt{Coherence}}) = \alpha + \beta_1 \text{\texttt{Narrativity}}$ and we apply epsilon smoothing for cases where coherence scores are exactly 0 or 1. We observe that there is generally a highly statistically significant association between increasing narrativity and increasing coherence.

\subsection{Narrative, Discourse, and Coherence}
Storytelling is pivotal in maintaining discourse-level coherence, a cognitive pattern that text linguistics identifies as a standard default assumption about any text or conversation until proven otherwise \cite{brown1983discourse}. When coherence is not immediately apparent, people actively search for evidence of its otherwise latent presence. In conversation scenarios, for example, the search for latent coherence can take the form of a cooperative effort between participants, where they collectively work towards making sense of the subject at hand \cite{grice1975logic}. 

This latent presence of coherence typically exists in textual spaces ``where it is not evident in the surface lexical or propositional cohesion'' \cite[179]{stubbs1983discourse}, such as implicit semantic relationships like the inclusion of cultural subtexts and schemata that underpin discursive structures \cite{brown1983discourse}, which are, in turn, often represented and conveyed in the form of narratives. Stories, therefore, serve as a \textit{vital communicative and cognitive resource for sense-making}, assisting individuals in navigating complex social and cultural contexts by providing a heuristic framework for understanding implicit meanings and relationships within discourse.

\subsubsection{Narratives Help Us Articulate and Understand Complex Arguments}
The narrative paradigm (NP) is a communication framework that posits that meaningful communication primarily occurs via forms of storytelling rather than discursive argumentation \cite{fisher1984narration}, based on the widely validated premise that the rhetorical logic of compelling narratives tends to be more persuasive than that of structured arguments \cite{dahlstrom2010role,schreiner2018argument,oschatz2020long}, particularly in the context of healthcare \cite{chen2016narrator,ballard2021impact}. NP construes humans as especially disposed to storytelling and whose communication channels and internal world models are inherently sensitive to and ordered by narratives. This notion is echoed by foundational works in psychology \cite{sacks1985man, mercier2017enigma}, philosophy of identity \cite{taylor1992sources}, and philosophy of literature \cite{bruner1987life,dennett1988everyone}. Moreover, NP introduces the binomial nomenclature ``homo narrans'' to highlight the importance of storytelling in the human condition, underscoring its essential role in ensuring meaningful communication, mutual intelligibility, and the formation of common sense \cite{fisher1984narration}.

As a model of human behavior, NP measures the effectiveness of conveying and understanding complex information through the narrative merit of its underlying stories. The judgment of this merit is based on two criteria of narrative rationality: 1) narrative coherence, the internal consistency of a story that allows it to make sense in its own context, and 2) narrative fidelity, the degree to which a story aligns or resonates with the receiver's existing understanding of reality and experiences with other stories \cite{fisher2021human}. In other words, the coherence of a message depends less on the completeness of its discursive structures than the self-consistency of the stories around which its content is framed.

\subsubsection{Narratives Maintain the Consistency of Our Own Internal World Models}
Cognitive narratology is a branch of narrative theory that examines the ``nexus of narrative and mind'' \cite[137] {herman2009basic} through the cognitive underpinnings of storytelling practices. It considers narratives to be not merely cultural artifacts for literary interpretation but cognitive instruments that serve as an indispensable resource for navigating the world and scaffolding, compartmentalizing, and negotiating our experience \cite{dannenberg2008coincidence,herman2013storytelling}. The foundation of this approach rests on frameworks in cognitive linguistics positing that language and thought are fundamentally shaped by stories and so-called ``parabolic projections'' that map them onto experience \cite{turner1996literary}.

Leveraging the semantics of possible worlds \cite{Kripke1963-KRISCO} and theory of mind \cite{zunshine2006we}, cognitive narratologists argue that stories accomplish more than building worlds to ensure narrative fidelity with the interpreter's mental models; they also enhance the self-consistency and robustness of such mental models through the simulation of counterfactual scenarios \cite{lewis1979counterfactual} and alternative ``storyworlds'' \cite{ryan1991possible,herman2013storytelling,gerrig2018experiencing}. This exercise of make-believe not only facilitates narrative understanding but also enriches the interpreter's cognitive landscape by providing a dynamic platform for testing and refining mental models through imaginative engagement and adaptation \cite{dolevzel1998heterocosmica,ryan2001narrative}, a process that has been linked to evolutionary mechanisms \cite{easterlin2012biocultural}, especially a predisposition towards play in evolutionary psychology \cite{boyd2009origin}.

Earlier periods of AI research took a particular interest in this mind-narrative nexus due to narrative's capacity to facilitate complex interpretations with very limited discourse-level information available. Computer scientists have attempted to represent these capabilities formally as 1) cognitive mechanisms like story grammar that could parse stories into sequences and embeddings \cite{mandler2014stories}, 2) frame representations of commonsense knowledge that reduce interpretive complexity by schematizing expected sequences of events and storing them in memory \cite{schank2013scripts}. In addition, cognitive narratologists have also suggested that the narrative semantics of possible worlds, in which the coherence of a narrative is complete as its ability to construct immersive storyworlds compellingly, could serve as a hermeneutic approach to probe into the neural processes of automatic story generation \cite{ryan1991possible}.

\subsubsection{Narratives Enable Patients to Negotiate the Coherence of Their Experiences}
One prominent display of narratives' cognitive affordances is in the medical domain. Narratives have been documented to play a critical role in emotional recovery and rehabilitation in the aftermath of traumatic events that have disrupted the internal coherence of patients' understanding of the world. Clinically validated narrative interventions have improved patient care and outcomes, especially by helping patients grapple with cataclysmic ruptures in the fabric of their experience \cite{rosario2018narrative, coats2020integration}. Therapeutically, narratives serve dual purposes: for patients, narratives provide a vital sense of agency, giving them control over how to tell their own stories, a process crucial for reconstructing a coherent identity and self-concept \cite{frank2013wounded}; for physicians, engaging with patient illness narratives allows them to enhance their empathy by gaining insights into the patient as a whole person, rather than viewing them merely as a collection of symptoms or statistics \cite{charon2001narrative}.

\section{Limitations and Directions for Future Research}

This paper advances a systematic defense of LLM confabulations as a potential resource instead of a categorically negative pitfall. We believe the mental picture that LLMs hallucinate because they are untrustworthy, unfaithful, and ultimately un-humanlike is an oversimplified one. Instead, they confabulate and exhibit narrative-rich behavioral patterns closely resembling the human urge of storytelling -- perhaps hallucinations make them more like us than we would like to admit.

While our current findings reveal intriguing associations between increased narrativity and significant increases in coherence, we must refrain from asserting that narrativity drives coherence. However, this perspective does garner substantial support from the interdisciplinary perspectives presented in our qualitative analysis. This paper serves as a preliminary exploration in this direction, but we require more robust methods for modeling narratives and more comprehensive human evaluations to elucidate the intricacies of this association. We are eager to implement these steps in the next phase of our study.

The textual behaviors and effects we observe in LLM hallucination benchmarks, i.e., higher degrees of narrativity and coherence and the significant correlation between them, are widely considered beneficial in human-to-human communication. But the extent to which these affordances generalize to human-AI interactions—whether humans indeed benefit more in terms of user experience from engaging with high-narrative confabulations versus low-narrative but factual generations—needs to be further validated with human-based evaluations. We plan to follow up this study with experiments with human participants to verify the benefits of narrative engagement as hypothesized. 

If the effectiveness of narratively rich confabulations is robustly validated, it could lead to exciting avenues for future research, testing the efficacy of confabulations across diverse domains. Beyond the applications we have already outlined in Section 1, exploring the utility of confabulations in fields such as journalism and advertising could yield valuable insights. By adopting confabulations as a conceptual framework, we could inspire more cross-disciplinary explorations, opening up new possibilities for LLM applications that extend beyond purely factual text generations.

\section{Acknowledgement}
For the research and completion of this paper, we acknowledge the resources and support provided by the McGill Faculty of Arts, McGill Department of English, and the Division of Research and Faculty Development at the Harvard Business School. Partial funding for the project was provided by Siva Reddy through a FRQNT New Researchers award and a NSERC Discovery grant. We thank Maria Antoniak for her help with the paper. We also acknowledge the helpful feedback from audiences at the University of Chicago’s Globus Labs, Data Science Institute, and Knowledge Lab, and Harvard Digital Data Design Institute. Finally, we sincerely thank our anonymous reviewers for their generous time, attention, and feedback.

\bibliography{acl_latex}
\end{document}